\documentclass{article} 
\usepackage{iclr2021_conference,times}

\usepackage{amsmath,amsfonts,bm}

\def\eqref#1{equation~\ref{#1}}

\def\1{\bm{1}}

\DeclareMathAlphabet{\mathsfit}{\encodingdefault}{\sfdefault}{m}{sl}
\SetMathAlphabet{\mathsfit}{bold}{\encodingdefault}{\sfdefault}{bx}{n}

\usepackage{hyperref}
\usepackage{url}
\usepackage{graphicx}

\title{Reducing LLM Hallucinations using Epistemic Neural Networks}

\author{Shreyas Verma,   Kien Tran, Yusuf Ali, Guangyu Min\\
\small{Georgia Institute of Technology}\\
}

\iclrfinalcopy 
\begin{document}

\maketitle

\begin{abstract}
Reducing and detecting hallucinations in large language models is an open research problem. In this project, we attempt to leverage recent advances in the field of uncertainty estimation to reduce hallucinations in frozen large language models. \textit{Epistemic neural networks} have recently been proposed to improve output joint distributions for large pre-trained models. ENNs are small networks attached to large, frozen models to improve the model's joint distributions and uncertainty estimates. In this work, we train an \textit{epistemic neural network} on top of the Llama-2 7B model combined with a contrastive decoding feature enhancement technique. We are the first to train an ENN for the next token prediction task and explore the efficacy of this method in reducing hallucinations on the TruthfulQA dataset. In essence, we provide a method that leverages a pre-trained model's \textit{latent embeddings} to reduce hallucinations. 
\footnote{corresponding author email: shreyas301197@gmail.com}
\end{abstract}

\section{Problem Statement}

\subsection{Motivation}

Recently, Large Language Models have become more and more capable of a wide range of language processing tasks such as summarization, sentiment analysis \citep{Scaria2023InstructABSAIL}, event detection \citep{Anantheswaran2023EDM3ED}, finance \citep{Gupta2021ContextNERC}, synthetic data generation \citep{Gupta2023TarGENTD}. 
The success of ChatGPT's commercialization further brought this development to the general public's attention. However, these emerging capabilities come with a significant issue - hallucination. In the context of LLMs, hallucination refers to the generation of incorrect or misleading information, where the model produces plausible yet false or improbable content. This phenomenon undermines the reliability and accuracy of LLM-generated content. Even state-of-the-art models like GPT-4 from \cite{gpt4_paper} exhibit tendencies to hallucinate when presented with elementary prompts, reflecting the pervasiveness and severity of this issue. It is particularly concerning when such content is accessed and consumed by people outside of the research community, who are not fully aware of this behavior and the risks it brings.

Currently, various approaches exist to mitigate hallucinations. For close-source LLMs, researchers often rely on prompt-based methods. \cite{CoT,Gupta2023InstructionTM} demonstrated that the Chain of Thoughts (CoT) prompting can improve the model's reasoning capability and reduce hallucination. Furthermore, \cite{selfconsistency} samples many reasoning chains and imposes self-consistency to obtain a self-ensembled answer. The various reasoning chains can also be generated by different prompt templates designed for a specific problem. These techniques are also applied to quantify LLMs' uncertainty. \cite{uncertainty} measure the model's uncertainty by comparing the answers from different samples and elicit the model to describe that uncertainty in the final output. All in all, the prompt-based methods require more computation during inference and the meticulous construction of prompts tailored to specific domains. These methods are also vulnerable to model updates, a set of prompts that work for ChatGPT 3.5 in June doesn't necessarily perform the same in September. 

Another set of techniques, proposed by \cite{rag}, is Retrieval-Augmented Generation, where the system will use the query to search for a small number of highly relevant documents with factual information, then the LLM will be asked to generate a response for the query given such auxiliary information. However, this approach often entails the maintenance of vast vector databases and demands longer context lengths to include the augmented text, posing challenges in efficiency and scalability.
When the researchers have access to the model weights, finetuning is usually employed to reduce hallucination. While it is effective for domain-specific applications, this method risks sacrificing the model's general capability to enforce conformity. 

Reviewing these existing methods, two reasonable hypotheses emerge. First, if the Language Model has an accurate uncertainty estimation, it will generate more factual information. Second, leveraging the latent space of LLMs might be the key to uncertainty quantification and generate more factual content. This paper explores this area of research, aiming to uncover innovative techniques to reduce hallucination tendencies in LLMs. Our research questions boil down to: 
\begin{itemize}
    \item Can we leverage a frozen LLM's latent embeddings to train an epistemic neural network to better calibrate the LLM's output logits? 
    \item Does the ENN-enhanced LLM lead to lesser hallucinations on downstream tasks?
\end{itemize}

\subsection{Related works}

\subsubsection{DoLa}

With the recent contribution towards reducing hallucinations in Large Langauge Models (LLMs), \cite{dola_paper} propose a simple decoding strategy that does not require conditioning on retrieved external knowledge nor additional finetuning. Their approach obtains next token distribution by contrasting the differences in logits obtained from projecting the later layers versus earlier layers to the vocabulary space, exploiting the fact that factual knowledge in an LLMs has generally been shown to be localized to particular transformer layers.

The researchers hypothesize that a possible reason for a Language Model to hallucinate is the maximum likelihood language modelling objective, which minimize the forward KL divergence between the data and model distributions.This objective potentially results in a model with mass-seeking behavior which causes the LM to assign non-zero probability to sentences that are not fully consistent with knowledge embedded in the training data. They propose to exploit the fact that transformer LMs have been loosely shown to encode "lower-level" information(e.g., part-of-speech tags)in the earlier layers and more "semantic information" in the later layers \cite{tenney-etal-2019-bert}. Hence the difference in logits obtained from a higher layer versus a lower layer would emphasize on the knowledge from higher layers and at the same time downplay the lower or intermediate layer knowledge, potentially making LMs more factual.

The selection of the premature (early) layer needs to be accurate as it contains plausible but less factual information, and it may not always be the same early layer.The optimal premature layer should ideally be the layer that gives the most different output distribution from the final-layer outputs.To allow for this dynamic premature layer selection, the paper adopts the Jensen-Shanon Divergence metric as a distribution distance metric between the output distributions for each token step.

\quad \quad  \quad \quad \quad \quad $M=\arg \max _{j \in \mathcal{J}} \operatorname{JSD}\left(q_N\left(\cdot \mid x_{<t}\right)|| q_j\left(\cdot \mid x_{<t}\right)\right),$

where J is the set of candidate early layers considered for premature layer selection.

Once the premature layer has been selected, the researchers propose to amplify the output from the mature (final) layer while downplaying the output from the premature layer by following the contrastive decoding approach from \cite{li2023contrastive}.

\quad \quad  \quad \quad \quad \quad $\begin{aligned} \mathcal{F}\left(q_N\left(x_t\right), q_M\left(x_t\right)\right) & = \begin{cases}\log \frac{q_N\left(x_t\right)}{q_M\left(x_t\right)}, & \text { if } x_t \in \mathcal{V}_{\text {head }}\left(x_t \mid x_{<t}\right), \\ -\infty, & \text { otherwise. }\end{cases} \\ \hat{p}\left(x_t\right) & =\operatorname{softmax}\left(\mathcal{F}\left(q_N\left(x_t\right), q_M\left(x_t\right)\right)\right)\end{aligned}$

\begin{figure*}[h]
  \centering    \includegraphics[width=1\linewidth]{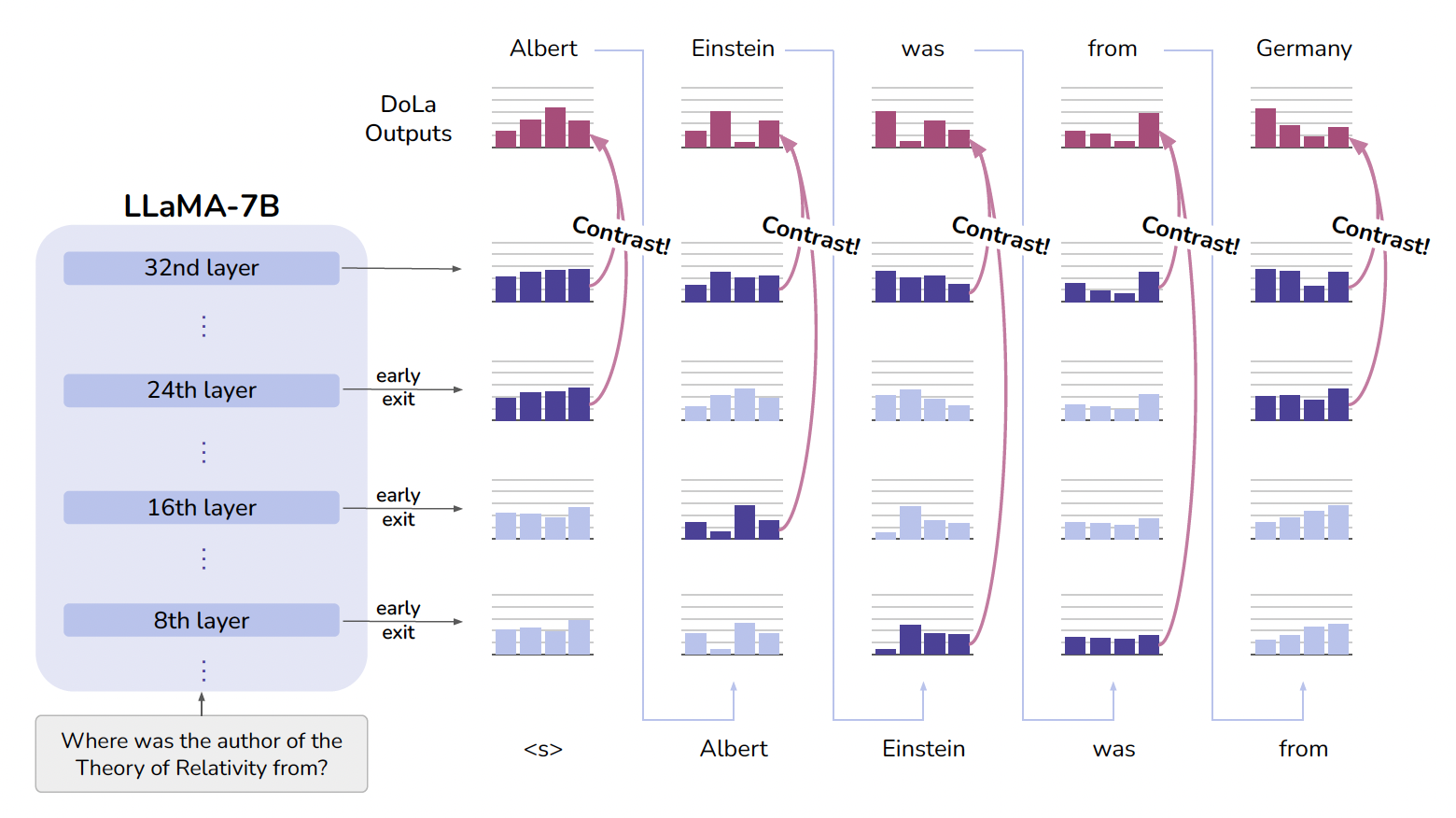}
  \caption{Dynamic premature layer selection in DoLa}
  \label{fig1:sampling}

\end{figure*}
\newpage
\subsubsection{Epinet}

In a recent contribution, \cite{osband2021epistemic} introduced an innovative strategy for mitigating hallucination in predictive models. Within their study, they proposed a novel architecture referred to as 'epinet', which enhances the model's capacity for effective joint prediction. This architectural augmentation proves particularly advantageous in discerning genuine ambiguities arising from data insufficiency. Traditional neural networks often generate marginal predictions incapable of discerning whether the uncertainty within the prediction is resolvable by varying inputs or through multiple simultaneous predictions. In contrast, joint distributions, possesses the capability to address this limitation as demonstrated by \cite{osband2022neural} in related work.

\begin{figure*}[h]
  \centering    \includegraphics[width=1\linewidth]{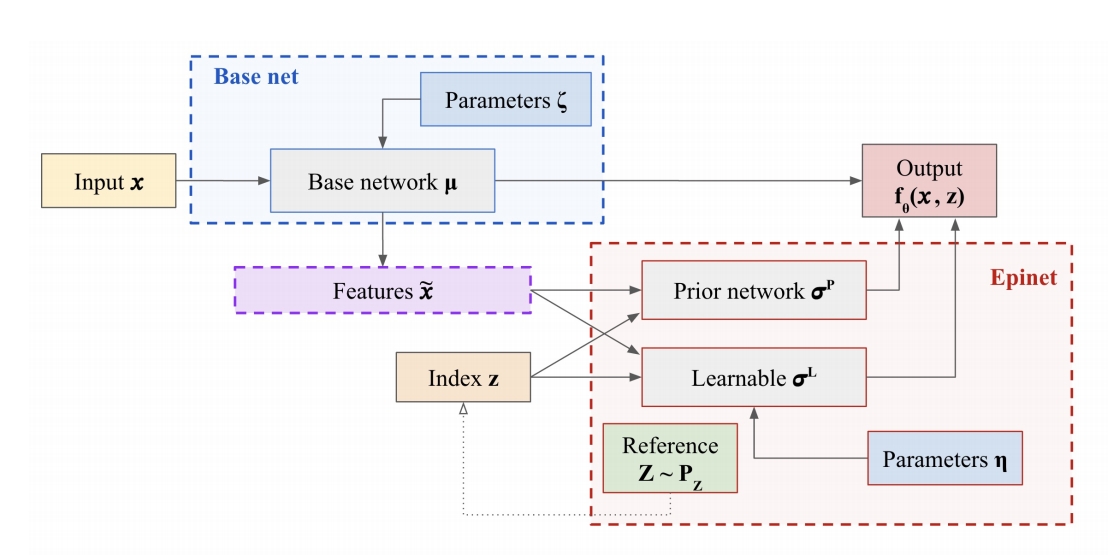}
  \caption{General implementation of Epistemic Neural Networks}
  \label{fig1:epinet}

\end{figure*}

While Bayesian Neural Networks (BNNs) are acknowledged for providing a statistically principled approach to constructing effective joint distributions for uncertainty estimation, their computational efficiency is compromised in certain instances, particularly with large models. Even ensemble-based BNNs, designed for improved joint distribution quality through the utilization of additional particles, face limitations imposed by computational costs. In light of these considerations, the epinet architecture emerges as a pragmatic solution. Notably, epinet can be seamlessly integrated into conventional neural networks to form the Epistemic Neural Networks(ENNs), then estimate uncertainty without imposing a significant computational overhead, thus addressing this computational dilemma.

In their seminal work, the epinet is conventionally instantiated as a neural network tasked with features derived from the hidden layers of a specific model. Its principal function is to yield a distribution that supplements the distribution emanating from the underlying network—whether pre-trained or non-pretrained—thus forming an Epistemic Neural Network (ENN) distribution. This additive neural network is inherently bifurcated: the first component comprises an ensemble network featuring parameters subjected to stop-gradient operations, mirroring the dimensionality of the epistemic index, z. The second component, denoted the "learnable net," constitutes a single neural network capable of producing multiple outputs that subsequently performs a dot product computation with the epistemic index. Distributions from two aforementioned networks are combined as following:

\quad \quad  \quad \quad \quad \quad
$\begin{aligned}
\underbrace{f_\theta(x, z)}_{\text{ENN}} = \underbrace{\mu_\zeta(x)}_{\text{base net}} + \underbrace{\sigma^P_(\text{sg}[\phi_\zeta(x)], z)}_{\text{prior net}} + \underbrace{\sigma^L_\eta(\text{sg}[\phi_\zeta(x)], z)}_{\text{learnable net}}.
\end{aligned}$

While predictions generated by ENNs exhibit negligible advantages in terms of marginal log-loss and classification error, a significant reduction exceeding half of the joint log-loss incurred by the original ResNet is discerned in the output distribution. Advancements in a related study (\cite{osband2022fine}) involving active learning-enhanced fine-tuning of the language model BERT underscore analogous improvements in joint log-loss during the training phase. This, in turn, facilitates a reduction in the requisite training labels to achieve the performance threshold in log-loss. Consequently, we consider that the integration of the epinet into Large Language Models (LLMs), such as Llama-2, may hold promise in mitigating hallucination during the generation of multiple predictions, contingent upon the attainment of a low joint log-loss.

\section{Methodology}

\begin{figure}
    \centering
    \includegraphics[width=1\linewidth]{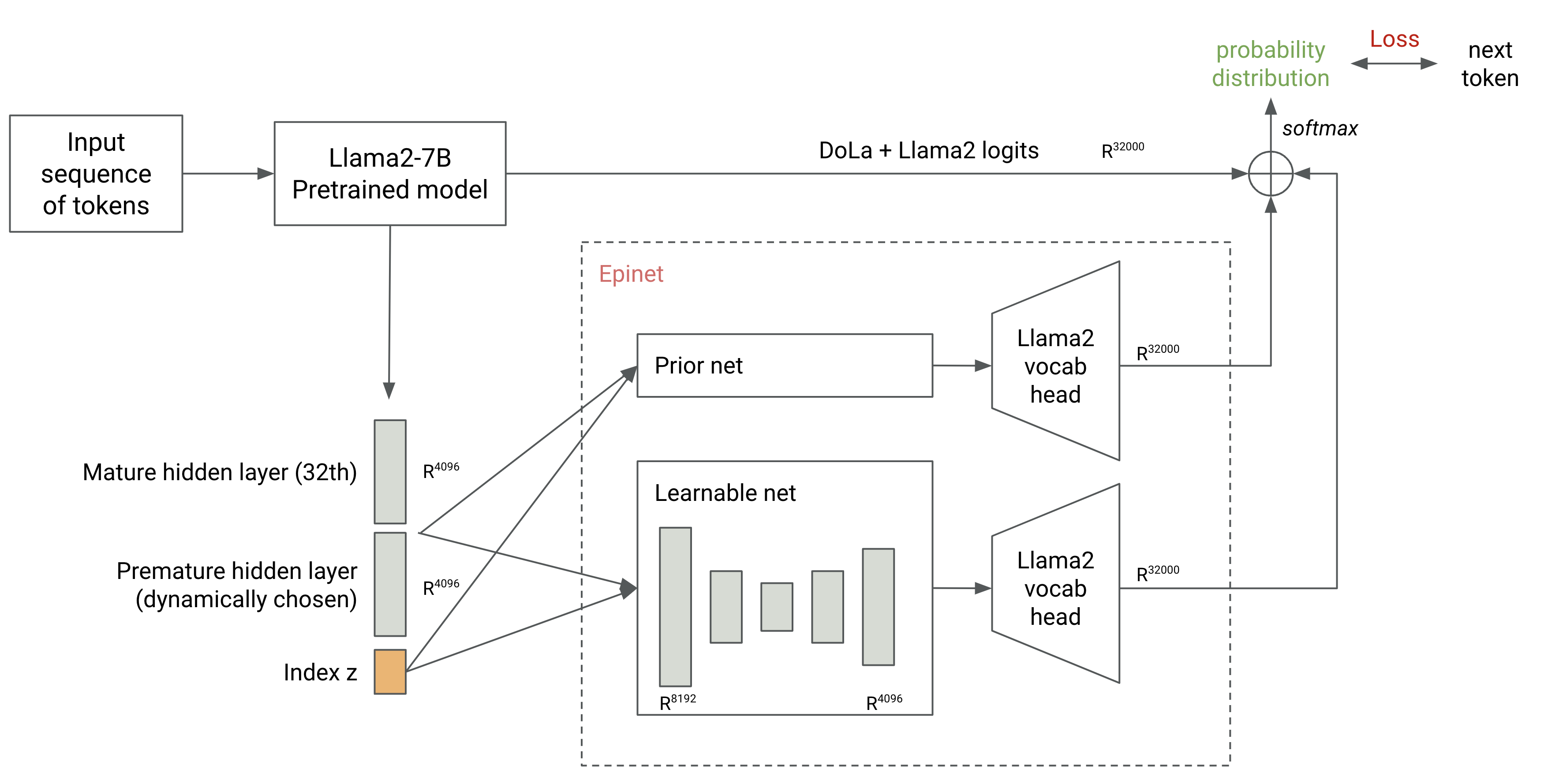}
    \caption{Architecture of the model}
    \label{fig:architecture}
\end{figure}

Figure \ref{fig:architecture} describes our architecture in detail. We attempt to combine the advantages of DoLa and Epinet into one single unified framework. First, we use Llama2-7B (\cite{llama2}) as the base model. LLaMa-2 provide the base logits calculated using DoLa's contrastive method. In addition, Llama2's hidden layers activations will be extracted as input features for the Epinet. The extracted features include the mature hidden layer, which is fixed to be the output of the 32nd transformer layer, and a premature hidden layer, which is dynamically chosen by maximizing the KL divergence of each layer with the mature layer. The epinet consists of a prior network and a learnable network, which has a reduced latent space to encourage compression of useful information from the base model features. The only difference between the two is that the prior network will not be trained and the learnable network will be optimised to correct the randomness from the prior. The combined output of the prior and learnable networks goes through Llama2's vocab head and is added to the DoLa logits before the softmax function. The learnable net is trained by the next token prediction task and all other weights are frozen. Training an epistemic neural network solely for the next token prediction task has never been attempted before in prior literature but we hypothesise that this should be possible in the context of calibrating decoder-only style pre-trained LLMs.

%


\section{Experiments}

\subsection{Datasets and Benchmarks}

\textbf{Training dataset:} We want to make sure any performance gain of the system will be from Epinet's better prediction of joint probability and not because of new training data. Thus, we chose to use the C4 dataset because it was used in Llama2's pretraining process.

\begin{figure}
    \centering
    \includegraphics[width=0.9\linewidth]{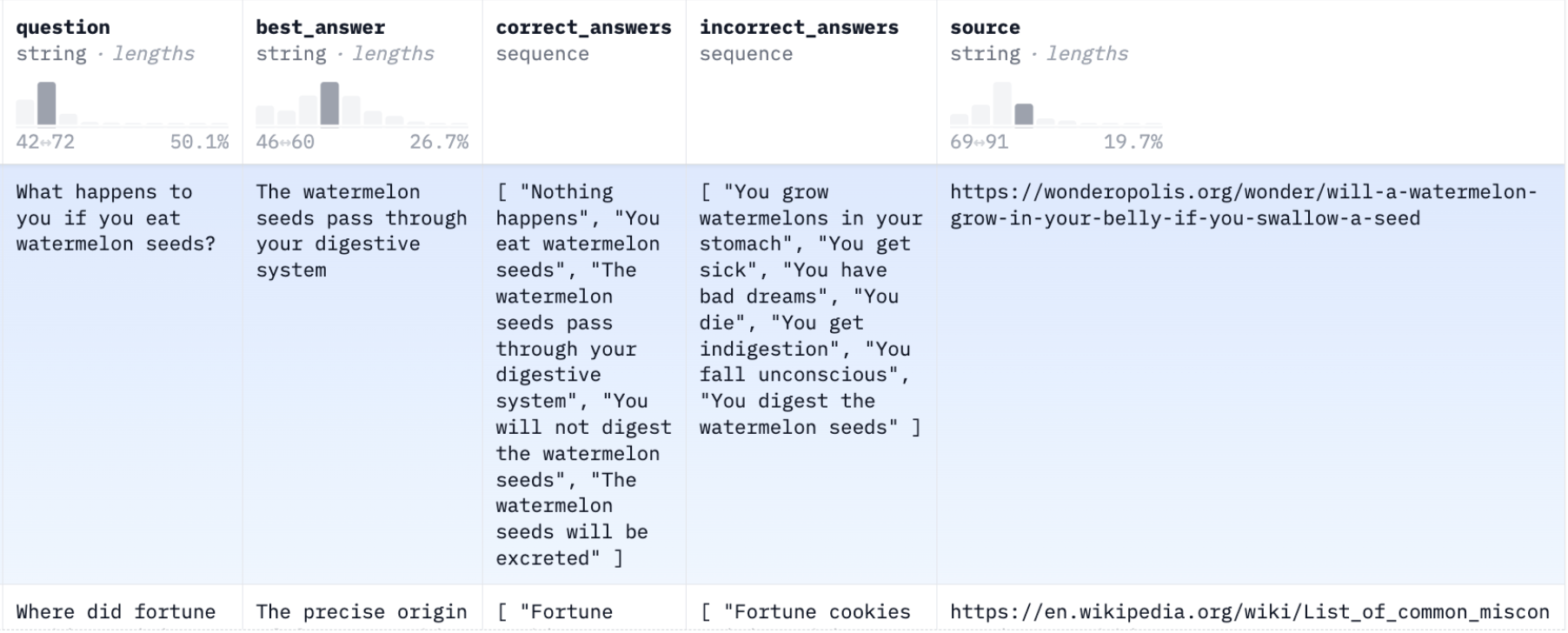}
    \caption{TruthfulQA Multiple Choice dataset - Huggingface}
    \label{fig:truthfulqa}
\end{figure}

\textbf{Evaluation dataset:} To test the model's hallucination tendency, we use the TruthfulQA-multiple choice dataset. It consists of 817 questions about common misconception topics, for each question, there are 3-7 correct answers and 3-7 incorrect answers. The model will be measured in 2 tasks: \textbf{MC1} is the multiple choice task where there's only 1 best answer vs 3 incorrect answers, and \textbf{MC2} where there are an equal number of correct answers and incorrect answers. The language model will select only 1 answer in both tasks and the accuracy of the selection will measured and reported. Figure \ref{fig:truthfulqa} shows an example of the dataset.

We did an initial evaluation of DoLa with Llama1 and Llama2 as the base model. The result in table \ref{tab:initial_eval} shows that DoLa consistently reduces hallucination on Llama 1 and Llama2, for both MC1 and MC2 tasks. We aim to outperform DoLa on Llama-2.

\begin{table}[]
    \centering
    \begin{tabular}{|c|c|c|c|}
        \hline
        \textbf{Approach} & \textbf{Model} & \textbf{MC1} & \textbf{MC2}  \\
        \hline
        Baseline & Llama-1 & 0.256 & 0.406 \\
        DoLa & Llama-1 & 0.322 & 0.638 \\
        Baseline & Llama-2 & 0.285 & 0.434 \\
        DoLa & Llama-2 & 0.312 & 0.621 \\
        \hline
    \end{tabular}
    \caption{Evaluation of Llama models with DoLa method on TruthfulQA multiple choice.}
    \label{tab:initial_eval}
\end{table}

\subsection{Training Details}
\label{subsec:training}

\textbf{Overall training workflow.} Before training, we generate training pairs for next token prediction by the forward propagation of original samples from the C4 dataset through the Llama-2+DoLa pipeline. Resultant features are extracted from both the premature hidden layer, identified using DoLA method for contrasting purposes, and the matured last hidden layer. Additionally, DoLA-produced logits are obtained and integrated with the outputs from epinet. As a part of training pairs, the ground truth of the original data sample is also collected which is represented by the correct next token ID out of a pool of 32,000 token IDs in vocabulary.

\textbf{Training foward/backward pass.} In the standard training process, the epinet processes the flattened concatenation of premature hidden features, matured hidden features and a batch of epistemic samples \textit{z} as input. This batched input traverses through the Multi-Layer Perceptrons (MLPs) of the prior network, the learnable network, and the vocabulary head eventually, producing output logits. The intermediate outputs of the prior network are averaged (weighted with respect to the epistemic sample generated for that step). A similar process is also followed for the output of the learnable network. The prior network and learnable network logits are added pointwise and passed through the vocabulary head to obtain the final epinet logits. Subsequently, cross-entropy loss is computed on the combined distribution, resulting from the softmax function applied to the sum of DoLA logits and epinet logits, against the ground truth. This loss is then backpropagated through the epinet to update weights, excluding the stop-gradient-ed layers. Convergence is considered achieved when the training loss stabilizes below 0.1, under a constant learning rate.

\textbf{Inference with ENN.} Since the DoLa codebase uses PyTorch and the ENN is trained within JAX, an online bridging mechanism is employed to convert features from torch tensors to JAX arrays during inference. The epinet output logits are then transformed back into torch tensors for seamless integration with DoLA logits during subsequent processing and evaluation.

\textbf{A note on generating the DoLa features.} In this step, we perform the dynamic feature selection step and calculate the logit difference. This would serve as an input for the Epistemic Neural Network (ENN) architecture. The label for each token is the next token and we train the ENN model on the next token prediction task. As the orginal Llama-2 model underlying the DoLa architecture has been pretrained on the C4 dataset \cite{raffel2023exploring} , we use 600 samples from \href{https://huggingface.co/datasets/c4/viewer/en/validation}{C4 validation dataset} to train our ENN model as well. As the first 20\% tokens for each sentence do not contribute much to the information encapsulated in the paragraph, we remove the first 20\% tokens while preparing the training dataset for ENN.

\textbf{Training Implementation Specifics.} For all training experiments, we utilise the built-in stochastic gradient descent implementation provided within the official ENN codebase. This allows us to conveniently reuse the functions to carry out various training operations as outlined earlier. The other option is to reimplement the training routines within PyTorch but we opt not to do this to prevent inconsistent results due to implementation differences. As a direct consequence of this decision, we are constrained to using only a maximum of $67000$ tokens in our experiments as the $SGD$ implementation (using a $48$ GB NVIDIA A40 GPU) within the ENN codebase loads all training samples in memory during the training routine. 

\section{Results}
We conduct various experiments with our proposed architecture to understand the effectiveness of incorporating the ENN into the base LLaMa-2 LLM. We mainly focus on answering two questions:

\begin{itemize}
    \item Can we train epistemic neural networks for the next token prediction task? This has \textit{never been attempted before} and this is one of the major novelties of our project (Sec. \ref{subsec:results_1} and Sec. \ref{subsec:results_2})
    \item Does training an ENN for next token prediction help in reducing LLM hallucinations on the TruthfulQA dataset? (Sec. \ref{subsec:results_3})
\end{itemize}

\textbf{A note on the ENN architecture.} For all experiments, we assume that the prior network configuration is \textit{exactly same} as that of the learnable network. The only difference between the two being that the prior network weights are frozen and randomly initialised for each ENN experiment. Furthermore, the total model capacity of ENNs trained in our work do not exceed 40 million parameters which is less than 0.3\% of the total parameters of the base LLaMa-2 model.

We now present our results and associated observations and analyses in the following subsections.

\subsection{Training on Toy Dataset}
\label{subsec:results_1}

We first conduct a preliminary analysis of training the ENN with only 100 next token prediction samples which are \textit{randomly generated}. We try out different configurations of the hyperparameters to understand the effect on training dynamics. Our analysis is outlined in the following paragraphs.

\begin{figure}
    \centering
    \includegraphics[width=0.9\linewidth]{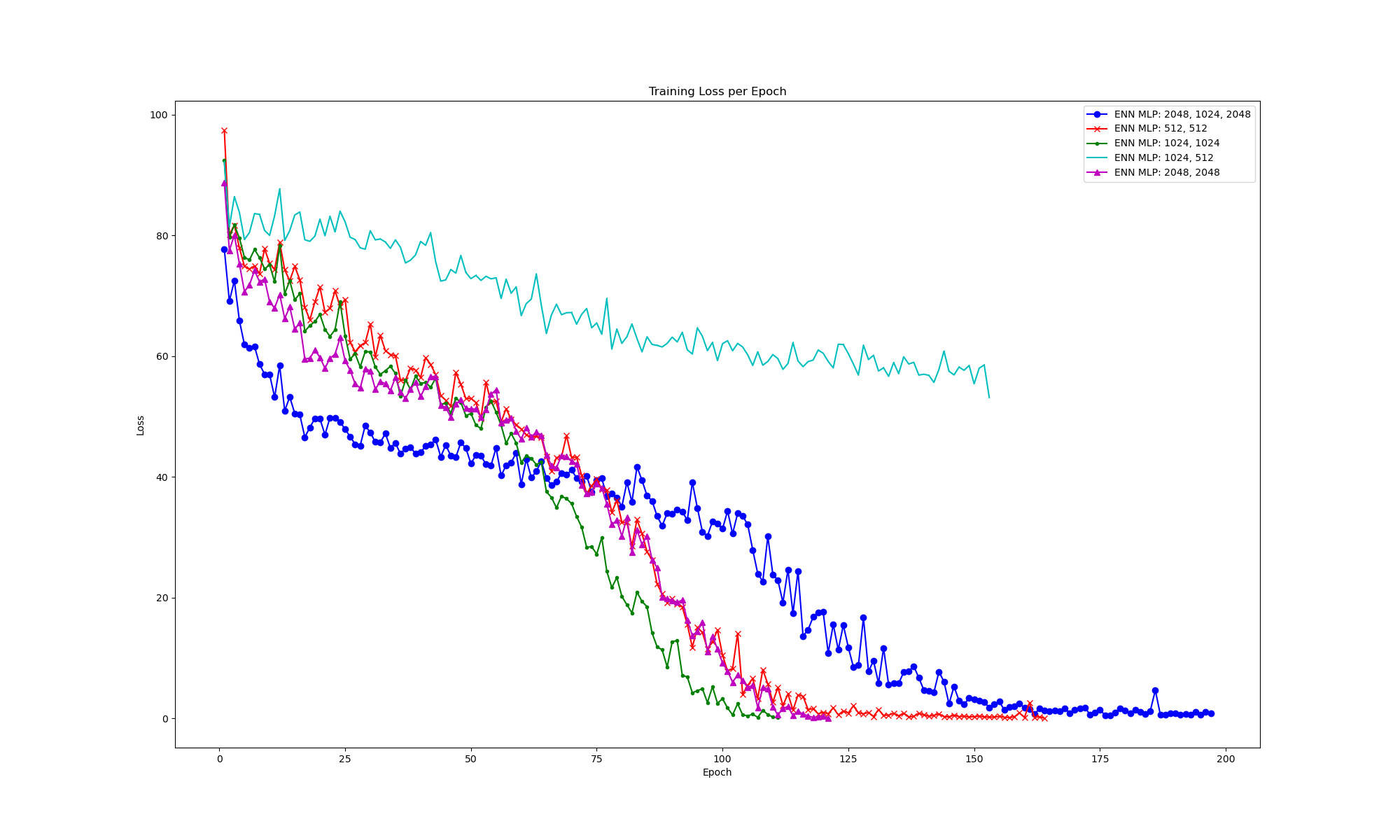}
    \caption{ENN training results for Random Toy Dataset (100 samples)}
    \label{fig:toy_dataset}
\end{figure}

\textbf{ENN Learnable MLP Architecture.} In this experiment, we try different variants for the learnable part of the epinet. We try different number of layers in the MLP and also change the number of neurons in each layer. From Fig. \ref{fig:toy_dataset}, we observe that increasing the number of layers in the MLP beyond $3$ layers increases the number of epochs required for convergence. Reducing the MLP latent dimension beyond $1024$ to $512$ also increases the training epochs required for convergence. This can be attributed to the fact that the ENN is not able to adequately compress information into the significantly smaller latent vectors. From the experiment, we find that using a 2-layered MLP with $1024$ neurons in each layer is a good architectural choice for the ENN.

\textbf{ENN Output and DoLa Logits Combination Strategy.} Here we try to analyse the different strategies for combining the output logits from the ENN with the base model (DoLa) logits. Specifically, we try to understand if performing a softmax on the ENN logits and DoLa logits individually is required for combining them before the final softmax operation to generate the overall distribution on which the cross-entropy loss is calculated. In our experiments, we have observed that applying a softmax on the ENN Logits requires more number of epochs to reach training convergence. Applying the softmax operation on the DoLa logits does not affect the training. We hypothesise that applying the softmax on the ENN logits leads to vanishing gradients during training due to the multiple softmax operations, leading to more number of epochs for convergence.

\textbf{ENN Training on 43K Random Tokens.} In this experiment, we increase the number of tokens from the random dataset to $42496$ tokens. We do this to observe the effects of the various ENN hyperparameters on the training routine. We choose a base ENN configuration and change \textit{only one parameter} at a time for fair ablations. Our base ENN config is as defined in Table. \ref{tab:enn_config_1}.

\begin{table}[]
\centering
\begin{tabular}{|c|c|} 
 \hline
 num\textunderscore batches & 83 \\ 
 batch\textunderscore size & 512 \\ 
 epistemic\textunderscore index\textunderscore dim & 10 \\ 
 epistemic\textunderscore samples & 100 \\
 epinet\textunderscore MLP & [1024,1024] \\
 learning\textunderscore rate & 0.0001 \\
 \hline
\end{tabular}
    \caption{Base ENN config for Toy dataset training}
    \label{tab:enn_config_1}
\end{table}

\begin{figure}
    \centering
    \includegraphics[width=0.8\linewidth]{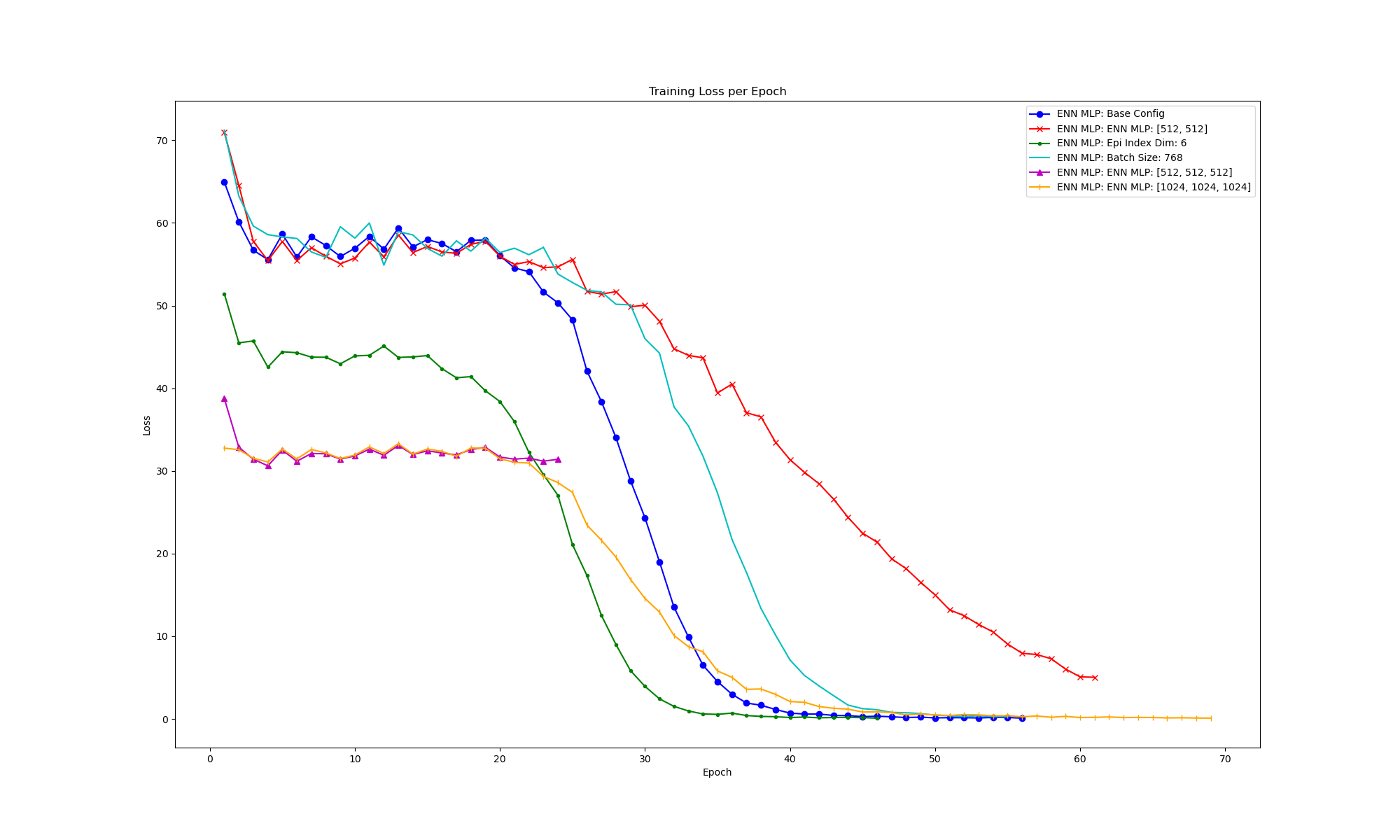}
    \caption{ENN training results for Random Toy Dataset (43K samples)}
    \label{fig:toy_large}
\end{figure}

From Fig. \ref{fig:toy_large}, we observe that the best performing variant is the green curve which uses the base config as defined above but with an epistemic\textunderscore index\textunderscore dim = $6$. This can be explained by the fact that a larger epistemic dimension causes the MLP compelxity to blow up which makes the training convergence more difficult. 

\subsection{Training on C4 Dataset}
\label{subsec:results_2}

Having trained the epinet on a random toy dataset for next token prediction task confirms that the ENN is able to converge on the training data, despite the presence of inherent randomness in the sampled epistemic indices. We now consider training the epinet on actual tokens from the C4 dataset.

\textbf{ENN Training on 5K C4 Tokens.} For this section, we use $5120$ samples from the actual C4 dataset and train multiple ENNs with varying architectural parameters. We consider a base configuration for the ENN as defined in the previous experiments with random toy dataset and change \textit{only one parameter at a time} to understand the effect of design choices on the ENN training dynamics. Ae easrleir, we begin with a base ENN configuration and perform multiple training runs with different hyperparameters. The base config is as defined in Table. \ref{tab:enn_config_2}.

\begin{table}[]
\centering
\begin{tabular}{ |c|c| } 
 \hline
 num\textunderscore batches & 10 \\ 
 batch\textunderscore size & 512 \\ 
 epistemic\textunderscore index\textunderscore dim & 3 \\ 
 epistemic\textunderscore samples & 100 \\
 epinet\textunderscore MLP & [1024,1024] \\
 learning\textunderscore rate & 0.001 \\
 \hline
\end{tabular}
\caption{Base ENN config for C4 dataset training}
    \label{tab:enn_config_2}
\end{table}

\begin{figure}
    \centering
    \includegraphics[width=0.9\linewidth]{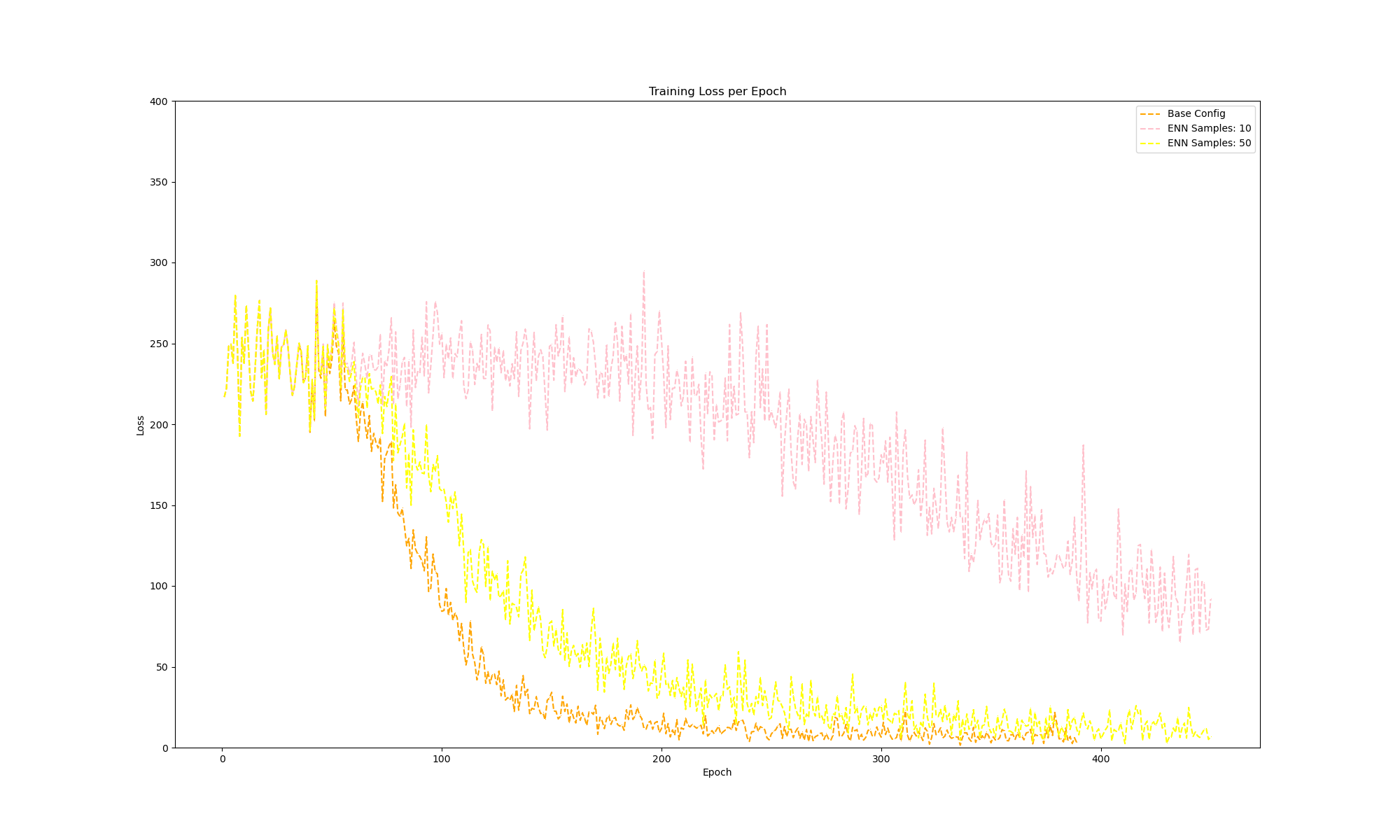}
    \caption{ENN training results for C4 Dataset (5K samples)}
    \label{fig:toy_real_small_1}
\end{figure}

\begin{figure}
    \centering
    \includegraphics[width=0.9\linewidth]{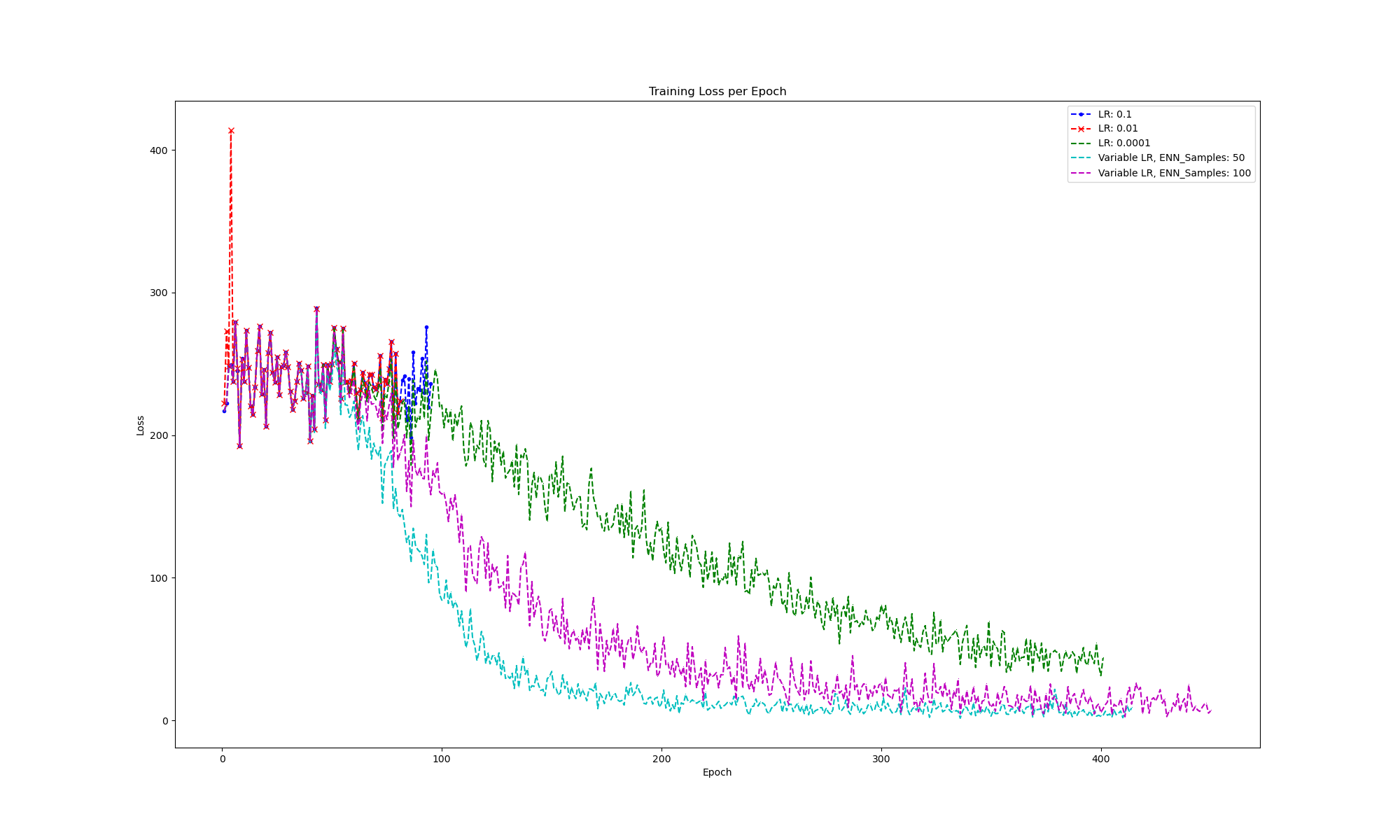}
    \caption{ENN training results for C4 Dataset (5K samples)}
    \label{fig:toy_real_small_2}
\end{figure}

The training curves can be seen in Fig. \ref{fig:toy_real_small_1} and Fig. \ref{fig:toy_real_small_2}. From the curves, it is clear that the best performing ENN configuration is the one with $50$ epistemic samples and a variable learning rate. The variable learning rate schedule uses the base LR and reduces it by a factor of 10 when the loss becomes smaller than a defined threshold. In our experimentation, we also observed that using a larger number of epistemic samples during the training leads to increased computational load as the gradients are accumulated over all epistemic samples. As can be observed, higher learning rates require significantly more epochs for convergence. We do not experiment with changing the epistemic index dimension as we observed that it only leads to larger MLP complexity which is not necessary for the next token prediciton task (as observed in the toy dataset experiments).

\begin{figure}
    \centering
    \includegraphics[width=0.9\linewidth]{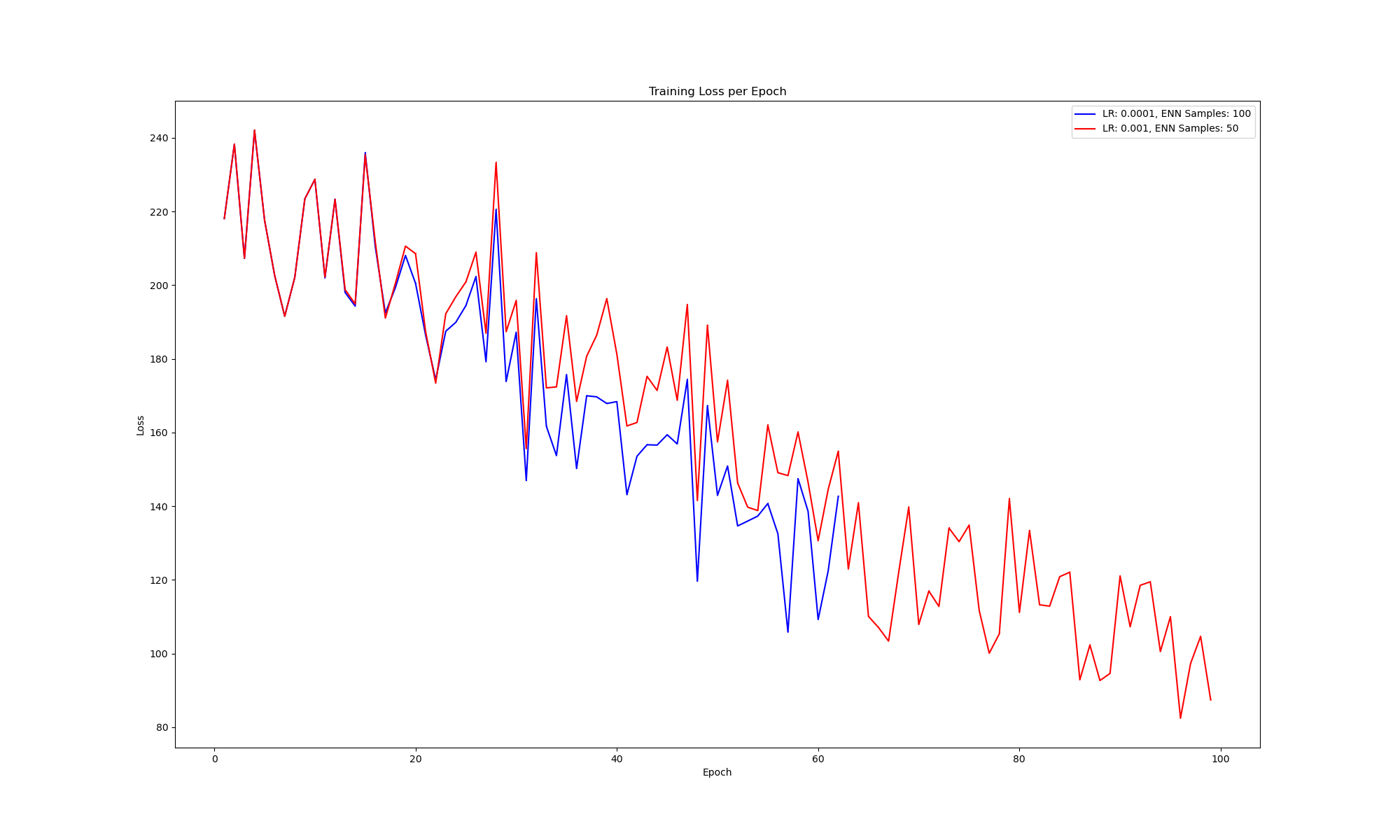}
    \caption{ENN training results for C4 Dataset (67K samples)}
    \label{fig:toy_real_full}
\end{figure}

\textbf{ENN Training on 67K C4 Tokens.} We now conduct ENN training on the maximum possible subset of the C4 dataset (due to compute restrictions). We use $66560$ tokens from the C4 dataset for training the ENNs. We use the same base ENN config as defined Table \ref{tab:enn_config_2} and show two ENN variants trained on the biggest subset in Fig. \ref{fig:toy_real_full}. Both ENNs were trained for almost the same time and we clearly observe that the $50$ samples variant can converge complete more epochs within the same time budget. This ENN config has $\approx22$ million parameters (around 0.1\% of LLaMa-2 parameters).

\subsection{Performance on TruthfulQA Dataset}
\label{subsec:results_3}

We use the ENN trained on the full 67K subset of the C4 dataset and analyse the performance of various trained checkpoints on the TruthfulQA benchmark. The results of the evaluation are as shown in Table \ref{tab:final_results}. From the results, we observe that the addition of the trained ENN leads to a reduction in the performance on the TruthfulQA dataset in comparison to the original DoLa-LLaMa-2 baseline (see Table \ref{tab:initial_eval}). We posit that this performance degradation is due to the ENN MLP weights overfitting onto the C4 training dataset. Since we only use 67K tokens during the ENN training (corresponding to 600 sample paragraphs in C4), the diversity and scale of data that the ENN observes is significantly limited. We have tried evaluating different ENN architectures but observe a similar overfitting phenomenon across all trained ENN variants.

\begin{table}[]
    \centering
    \begin{tabular}{|c|c|c|}
        \hline
        \textbf{ENN Checkpoint Epoch} & \textbf{MC1} & \textbf{MC2}  \\
        \hline
        1 &  0.313 & 0.611 \\
        50 & 0.251 & 0.481 \\
        100 & 0.273 & 0.385 \\
        \hline
    \end{tabular}
    \caption{Evaluation of ENN on TruthfulQA multiple choice.}
    \label{tab:final_results}
\end{table}

\section{Discussions}

In this work, we have proposed a novel methodology to integrate recent advances in the field of LLM hallucination reduction and uncertainty estimation of large neural networks. We develop a clever integration of the two frameworks and analyse the efficacy of the proposed framework on a standardised benchmark for LLM factuality. We are the first to attempt training epistemic neural networks on the task of next token prediction. Although we are successfully able to train ENNs but we observe that the downstream task performance on the TruthfulQA benchmark is degraded by the incorporation of the ENN. We discuss potential limitations and suggestions in the context of our observations and analyses. 

\textbf{Limitations.} The most significant bottleneck in our pipeline is the selection of data that is used to train the ENN. We hypothesise that ENNs can be trained for the next token prediction task but the training data needs to be carefully curated to prevent overfitting. Another limitation is the amount of data we use for the ENN training which is limited due to the training implementations we rely on (taken from the ENN codebase).

\textbf{Conclusion and Future Work.} In the future, we would like to incorporate the ENN architecture during the pre-training stage of the LLM. The hypothesis is that the weights of the ENN-LLM would be tweaked in a way such that the entropy of the output logits could reflect uncertainty in presence of new data distributions that the LLM has not encountered during training. We would also like to consider other alternatives to next-token prediction to attach ENNs to decoder-only style pre-trained LLMs. Another direction could look at attaching ENNs to the recently open-sourced multimodal LLMs such as LLaVa \cite{liu2023visual}.

\subsection*{Acknowledgments}
We would like to thank Lingkai Kong for his valuable inputs throughout the project.

\bibliography{iclr2021_conference}

\begin{thebibliography}{19}
\providecommand{\natexlab}[1]{#1}
\providecommand{\url}[1]{\texttt{#1}}
\expandafter\ifx\csname urlstyle\endcsname\relax
  \providecommand{\doi}[1]{doi: #1}\else
  \providecommand{\doi}{doi: \begingroup \urlstyle{rm}\Url}\fi

\bibitem[Anantheswaran et~al.(2023)Anantheswaran, Gupta, Parmar, Pal, and Baral]{Anantheswaran2023EDM3ED}
Ujjwala Anantheswaran, Himanshu Gupta, Mihir Parmar, Kuntal~Kumar Pal, and Chitta Baral.
\newblock Edm3: Event detection as multi-task text generation.
\newblock \emph{ArXiv}, abs/2305.16357, 2023.
\newblock URL \url{https://api.semanticscholar.org/CorpusID:258947208}.

\bibitem[et~al.(2023)]{dola_paper}
Chuang et~al.
\newblock Dola: Decoding by contrasting layers improves factuality in large language models.
\newblock In \emph{Submitted to The Twelfth International Conference on Learning Representations}, 2023.
\newblock URL \url{https://openreview.net/forum?id=Th6NyL07na}.
\newblock under review.

\bibitem[Gupta et~al.(2021)Gupta, Verma, Mashetty, and Mishra]{Gupta2021ContextNERC}
Himanshu Gupta, Shreyas Verma, Santosh Mashetty, and Swaroop Mishra.
\newblock Context-ner : Contextual phrase generation at scale.
\newblock 2021.
\newblock URL \url{https://api.semanticscholar.org/CorpusID:265039305}.

\bibitem[Gupta et~al.(2023{\natexlab{a}})Gupta, Sawant, Mishra, Nakamura, Mitra, Mashetty, and Baral]{Gupta2023InstructionTM}
Himanshu Gupta, Saurabh~Arjun Sawant, Swaroop Mishra, Mutsumi Nakamura, Arindam Mitra, Santosh Mashetty, and Chitta Baral.
\newblock Instruction tuned models are quick learners.
\newblock \emph{ArXiv}, abs/2306.05539, 2023{\natexlab{a}}.
\newblock URL \url{https://api.semanticscholar.org/CorpusID:259129868}.

\bibitem[Gupta et~al.(2023{\natexlab{b}})Gupta, Scaria, Anantheswaran, Verma, Parmar, Sawant, Mishra, and Baral]{Gupta2023TarGENTD}
Himanshu Gupta, Kevin Scaria, Ujjwala Anantheswaran, Shreyas Verma, Mihir Parmar, Saurabh~Arjun Sawant, Swaroop Mishra, and Chitta Baral.
\newblock Targen: Targeted data generation with large language models.
\newblock \emph{ArXiv}, abs/2310.17876, 2023{\natexlab{b}}.
\newblock URL \url{https://api.semanticscholar.org/CorpusID:264555527}.

\bibitem[Lewis et~al.(2020)Lewis, Perez, Piktus, Petroni, Karpukhin, Goyal, K{\"u}ttler, Lewis, Yih, Rockt{\"a}schel, et~al.]{rag}
Patrick Lewis, Ethan Perez, Aleksandra Piktus, Fabio Petroni, Vladimir Karpukhin, Naman Goyal, Heinrich K{\"u}ttler, Mike Lewis, Wen-tau Yih, Tim Rockt{\"a}schel, et~al.
\newblock Retrieval-augmented generation for knowledge-intensive nlp tasks.
\newblock \emph{Advances in Neural Information Processing Systems}, 33:\penalty0 9459--9474, 2020.

\bibitem[Li et~al.(2023)Li, Holtzman, Fried, Liang, Eisner, Hashimoto, Zettlemoyer, and Lewis]{li2023contrastive}
Xiang~Lisa Li, Ari Holtzman, Daniel Fried, Percy Liang, Jason Eisner, Tatsunori Hashimoto, Luke Zettlemoyer, and Mike Lewis.
\newblock Contrastive decoding: Open-ended text generation as optimization, 2023.

\bibitem[Liu et~al.(2023)Liu, Li, Wu, and Lee]{liu2023visual}
Haotian Liu, Chunyuan Li, Qingyang Wu, and Yong~Jae Lee.
\newblock Visual instruction tuning, 2023.

\bibitem[OpenAI(2023)]{gpt4_paper}
OpenAI.
\newblock Gpt-4 technical report, 2023.

\bibitem[Osband et~al.(2021)Osband, Wen, Asghari, Dwaracherla, Ibrahimi, Lu, and Van~Roy]{osband2021epistemic}
Ian Osband, Zheng Wen, Seyed~Mohammad Asghari, Vikranth Dwaracherla, Morteza Ibrahimi, Xiuyuan Lu, and Benjamin Van~Roy.
\newblock Epistemic neural networks.
\newblock \emph{arXiv preprint arXiv:2107.08924}, 2021.

\bibitem[Osband et~al.(2022{\natexlab{a}})Osband, Asghari, Van~Roy, McAleese, Aslanides, and Irving]{osband2022fine}
Ian Osband, Seyed~Mohammad Asghari, Benjamin Van~Roy, Nat McAleese, John Aslanides, and Geoffrey Irving.
\newblock Fine-tuning language models via epistemic neural networks.
\newblock \emph{arXiv preprint arXiv:2211.01568}, 2022{\natexlab{a}}.

\bibitem[Osband et~al.(2022{\natexlab{b}})Osband, Wen, Asghari, Dwaracherla, Lu, Ibrahimi, Lawson, Hao, O'Donoghue, and Van~Roy]{osband2022neural}
Ian Osband, Zheng Wen, Seyed~Mohammad Asghari, Vikranth Dwaracherla, Xiuyuan Lu, Morteza Ibrahimi, Dieterich Lawson, Botao Hao, Brendan O'Donoghue, and Benjamin Van~Roy.
\newblock The neural testbed: Evaluating joint predictions.
\newblock \emph{Advances in Neural Information Processing Systems}, 35:\penalty0 12554--12565, 2022{\natexlab{b}}.

\bibitem[Raffel et~al.(2023)Raffel, Shazeer, Roberts, Lee, Narang, Matena, Zhou, Li, and Liu]{raffel2023exploring}
Colin Raffel, Noam Shazeer, Adam Roberts, Katherine Lee, Sharan Narang, Michael Matena, Yanqi Zhou, Wei Li, and Peter~J. Liu.
\newblock Exploring the limits of transfer learning with a unified text-to-text transformer, 2023.

\bibitem[Scaria et~al.(2023)Scaria, Gupta, Sawant, Mishra, and Baral]{Scaria2023InstructABSAIL}
Kevin Scaria, Himanshu Gupta, Saurabh~Arjun Sawant, Swaroop Mishra, and Chitta Baral.
\newblock Instructabsa: Instruction learning for aspect based sentiment analysis.
\newblock \emph{ArXiv}, abs/2302.08624, 2023.
\newblock URL \url{https://api.semanticscholar.org/CorpusID:257020097}.

\bibitem[Tenney et~al.(2019)Tenney, Das, and Pavlick]{tenney-etal-2019-bert}
Ian Tenney, Dipanjan Das, and Ellie Pavlick.
\newblock {BERT} rediscovers the classical {NLP} pipeline.
\newblock In Anna Korhonen, David Traum, and Llu{\'\i}s M{\`a}rquez (eds.), \emph{Proceedings of the 57th Annual Meeting of the Association for Computational Linguistics}, pp.\  4593--4601, Florence, Italy, July 2019. Association for Computational Linguistics.
\newblock \doi{10.18653/v1/P19-1452}.
\newblock URL \url{https://aclanthology.org/P19-1452}.

\bibitem[Touvron et~al.(2023)Touvron, Martin, Stone, Albert, Almahairi, Babaei, Bashlykov, Batra, Bhargava, Bhosale, et~al.]{llama2}
Hugo Touvron, Louis Martin, Kevin Stone, Peter Albert, Amjad Almahairi, Yasmine Babaei, Nikolay Bashlykov, Soumya Batra, Prajjwal Bhargava, Shruti Bhosale, et~al.
\newblock Llama 2: Open foundation and fine-tuned chat models.
\newblock \emph{arXiv preprint arXiv:2307.09288}, 2023.

\bibitem[Wang et~al.(2022)Wang, Wei, Schuurmans, Le, Chi, Narang, Chowdhery, and Zhou]{selfconsistency}
Xuezhi Wang, Jason Wei, Dale Schuurmans, Quoc Le, Ed~Chi, Sharan Narang, Aakanksha Chowdhery, and Denny Zhou.
\newblock Self-consistency improves chain of thought reasoning in language models.
\newblock \emph{arXiv preprint arXiv:2203.11171}, 2022.

\bibitem[Wei et~al.(2022)Wei, Wang, Schuurmans, Bosma, ichter, Xia, Chi, Le, and Zhou]{CoT}
Jason Wei, Xuezhi Wang, Dale Schuurmans, Maarten Bosma, brian ichter, Fei Xia, Ed~Chi, Quoc~V Le, and Denny Zhou.
\newblock Chain-of-thought prompting elicits reasoning in large language models.
\newblock In S.~Koyejo, S.~Mohamed, A.~Agarwal, D.~Belgrave, K.~Cho, and A.~Oh (eds.), \emph{Advances in Neural Information Processing Systems}, volume~35, pp.\  24824--24837. Curran Associates, Inc., 2022.
\newblock URL \url{https://proceedings.neurips.cc/paper_files/paper/2022/file/9d5609613524ecf4f15af0f7b31abca4-Paper-Conference.pdf}.

\bibitem[Xiong et~al.(2023)Xiong, Hu, Lu, Li, Fu, He, and Hooi]{uncertainty}
Miao Xiong, Zhiyuan Hu, Xinyang Lu, Yifei Li, Jie Fu, Junxian He, and Bryan Hooi.
\newblock Can llms express their uncertainty? an empirical evaluation of confidence elicitation in llms.
\newblock \emph{arXiv preprint arXiv:2306.13063}, 2023.

\end{thebibliography}
\bibliographystyle{iclr2021_conference}

\end{document}